%% file: arxiv.tex
\documentclass[10pt,twocolumn,letterpaper]{article}

\usepackage{cvpr}              
\usepackage{xcolor}

\usepackage[linesnumbered,ruled,vlined]{algorithm2e}
\usepackage{comment}
\usepackage{bm}


\definecolor{cvprblue}{rgb}{0.21,0.49,0.74}
\usepackage[pagebackref,breaklinks,colorlinks,allcolors=cvprblue]{hyperref}

\newcommand{\OURS}{PrEditor3D}

\SetKwInput{KwInput}{Input}                %
\SetKwInput{KwOutput}{Output}

\title{\OURS{}: Fast and Precise 3D Shape Editing}

\author{Ziya Erko\c{c}\textsuperscript{1}
\and
Can G\"{u}meli\textsuperscript{1}
\and
Chaoyang Wang\textsuperscript{2}
\and
Matthias Nie{\ss}ner\textsuperscript{1}
\and
Angela Dai\textsuperscript{1}
\and
Peter Wonka\textsuperscript{2,3}
\and
Hsin-Ying Lee\textsuperscript{2}
\and
Peiye Zhuang\textsuperscript{2}
\and
\textsuperscript{1}Technical University of Munich \, \textsuperscript{2}Snap Inc \,
\textsuperscript{3}King Abdullah University of Science and Technology\\ \\
\url{https://ziyaerkoc.com/preditor3d}
}

\begin{document}

\input{figure_tex/teaser}

\maketitle

\input{sec/0_abstract}
\input{sec/1_intro}

\input{sec/2_related_hy}

\input{sec/3_method}
\input{sec/4_results}

\input{sec/5_conclusion}

\clearpage
{
    \small
    \bibliographystyle{ieeenat_fullname}
    \bibliography{main}
}

\clearpage
\input{sec/X_suppl}

\end{document}

%% file: figure_tex/teaser.tex
\twocolumn[{%
	\renewcommand\twocolumn[1][]{#1}%
	\maketitle
        \vspace{-8mm}
	\begin{center}
            \captionsetup{type=figure}
		\includegraphics[width=\linewidth]{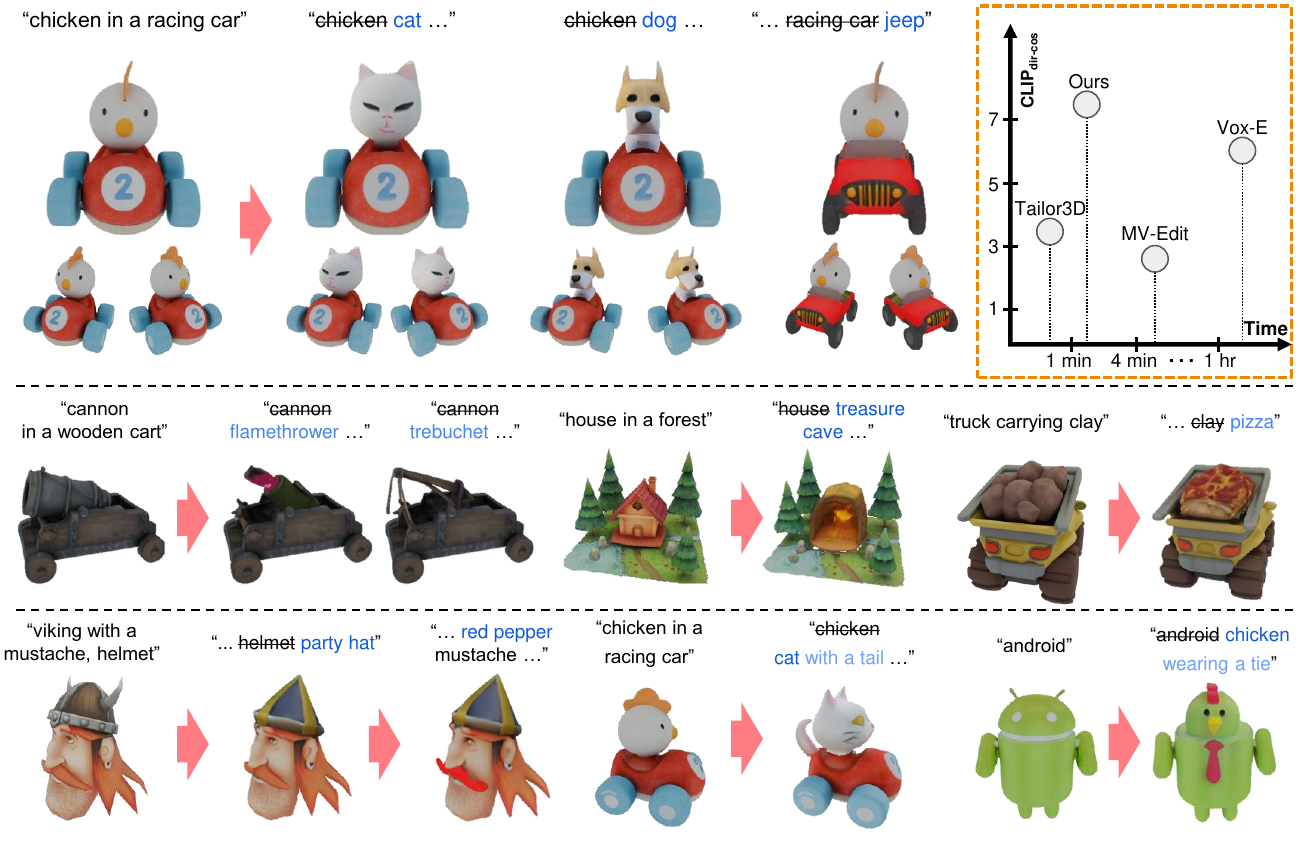}
        \vspace{-9mm}
		\captionof{figure}{
        \textbf{\OURS{}} is a (\textit{top}) fast and high-quality editing method that can perform precise and consistent editing only in the intended regions, keeping the rest identical.
        (\textit{mid}) It can handle diverse editing prompts with any given 3D object.
        (\textit{bottom}) Furthermore, it can support iterative editing, facilitating artistic workflow, and can also support editing multiple regions in a single run.
        }
		\label{fig:teaser}
	\end{center}    
}]

%% file: sec/0_abstract.tex
\begin{abstract}

We propose a training-free approach to 3D editing that enables the editing of a single shape within a few minutes. 
The edited 3D mesh aligns well with the prompts, and remains identical for regions that are not intended to be altered.
To this end, we first project the 3D object onto 4-view images and perform synchronized multi-view image editing along with user-guided text prompts and user-provided rough masks.
However, the targeted regions to be edited are ambiguous due to projection from 3D to 2D. 
To ensure precise editing only in intended regions, we develop a 3D segmentation pipeline that detects edited areas in 3D space, followed by a merging algorithm to seamlessly integrate edited 3D regions with the original input. Extensive experiments demonstrate the superiority of our method over previous approaches, enabling fast, high-quality editing while preserving unintended regions.
\end{abstract}

%% file: sec/1_intro.tex
\vspace{-.7cm}

\section{Introduction}
\label{sec:intro}

Recent 3D diffusion models can generate high-quality assets that closely align with the text prompts in the form of neural fields~\cite{mvdream, magic3d}, meshes~\cite{polydiff,meshgpt}, or Gaussian point clouds~\cite{gaussiandreamer}. Although these methods generate impressive results, they lack the essential capability for precise and controllable editing of the generated outputs, a critical requirement for iterative artistic workflows.
Effective 3D editing demands: (1) it should be fast enough to provide quick feedback,  ideally comparable to fast 3D generation algorithms, and (2) it must allow for precise local control, enabling users to keep specific parts of the model unchanged. 

Enabling precise and controllable editing is still an open challenge. Several initial approaches have been proposed to tackle the challenge of 3D editing \cite{voxe, instructnerf2nerf, shapeditor, tailor3d, interactive3d, mvedit}, providing promising results but suffering from slow runtime, lack of precise control, and/or lack of 3D consistency and quality.
Optimization-based techniques like SDS~\cite{voxe, shapeditor, interactive3d} or multiview training dataset updates~\cite{instructnerf2nerf} are computationally expensive, making interactive editing out of reach.

Additionally, they offer limited control over specific parts of the shape, as text prompts alone cannot precisely localize regions to be edited~\cite{voxe, instructnerf2nerf, shapeditor, tailor3d, mvedit}.
While Vox-E~\cite{voxe} and Shap-Editor ~\cite{shapeditor} propose a mechanism to prevent original parts of the shape from being altered during editing, they do not enable precise editing due to having only text as input.
Finally, one can observe various visual quality problems, such as the Janus problem, blurring, over-saturation, and overemphasizing texture changes while leaving the geometry intact or degrading.

To address these challenges, we propose a novel editing pipeline for 3D assets that is faster, more precise, and delivers high-quality results (See Fig.~\ref{fig:teaser}). 
As our primary goal is faster editing, we propose an editing framework leveraging a pipeline that consists of two components: a multi-view diffusion algorithm and a feed-forward mesh reconstruction.
Multi-view diffusion models can leverage superior 2D editing techniques, and the feed-forward mesh reconstruction bridges the gap between 2D and 3D.
For better controllability, we extend multi-view image generation to multi-view image editing using 2D masks to constrain the edits to user-specified regions. 
The 2D masks can take various forms, including manually selected regions, hand-brushed areas, or automatically generated segmentations.
We adopt DDPM inversion~\cite{editfriendlyddpm} to extract initial noise vectors from input multi-view images and execute Prompt-to-Prompt~\cite{prompt2prompt} on a multi-view diffusion model~\cite{mvdream}. 
We use 2D user-provided masks to blend edited and original views during the denoising.
However, due to the inherent ambiguity caused by projection from 3D to 2D, we cannot obtain ideal intended regions in 2D regardless of the granularity of the masks, as shown in Fig.~\ref{fig:occlusion}.
However, the masks are often too rough to precisely capture the intended semantic editing regions.
The masks are either too coarse so the unintended regions will be changed, or too fine-grained to allow reasonable editing.
Without additional spatial information in 3D, multi-view editing approaches cannot fully address this challenge. 
Therefore, simply adopting a feed-forward reconstruction method~\cite{gtr} to convert edited multi-views into a 3D mesh often leads to undesirable results.

To tackle this issue, we propose using the original 3D input and 3D segmentation. We first detect the intended editing region using Grounding DINO \cite{groundingdino} and SAM 2 \cite{sam} with the user mask and prompts. This gives an initial 2D segmentation that we subsequently lift to 3D. For this purpose, we use color coding in a multi-view to 3D reconstruction pipeline, named GTR~\cite{gtr}, to end up with a 3D segmentation that we can use during merging. Specifically, we paint 2D segmentations in a specific color, (e.g., green) and after reconstruction, we can detect which regions are edited by querying the color in the 3D field.
 
Then, we perform merging to maintain the original parts of the shape. 
We use 3D masks to detect edited/replaced parts in GTR's~\cite{gtr} voxel-feature space. We extract the edited part from the new shape and replace it with the old part from the original shape. That way, we guarantee that the remaining shape will remain identical. We apply a final average blending operation so that new parts and the original shape blend smoothly.

\input{figure_tex/occlusion}
\input{figure_tex/overview}
In summary, we make the following contributions:
\begin{itemize}
\item We propose a novel method for diffusion-based 3D object editing that is faster than previous work and enables precise, interactive editing.
\item The proposed method consists of user-guided and multi-view-synchronized editing and a feed-forward 3D reconstruction, enabling fast editing in a feed-forward manner.
\item To enable precise editing only to intended regions, we propose a voxel-based 3D segmentation method that utilizes multi-view segmentation information and propagates it to 3D, followed by an average blending operation to merge the edited and original objects.
\item Our editing method has superior quality to previous work. We show significant improvements in GPTEval3D~\cite{gpteval3d}, directional CLIP metrics, and extensive user studies. 
\end{itemize}

%% file: figure_tex/occlusion.tex
\begin{figure}
    \centering
    \includegraphics[width=\linewidth]{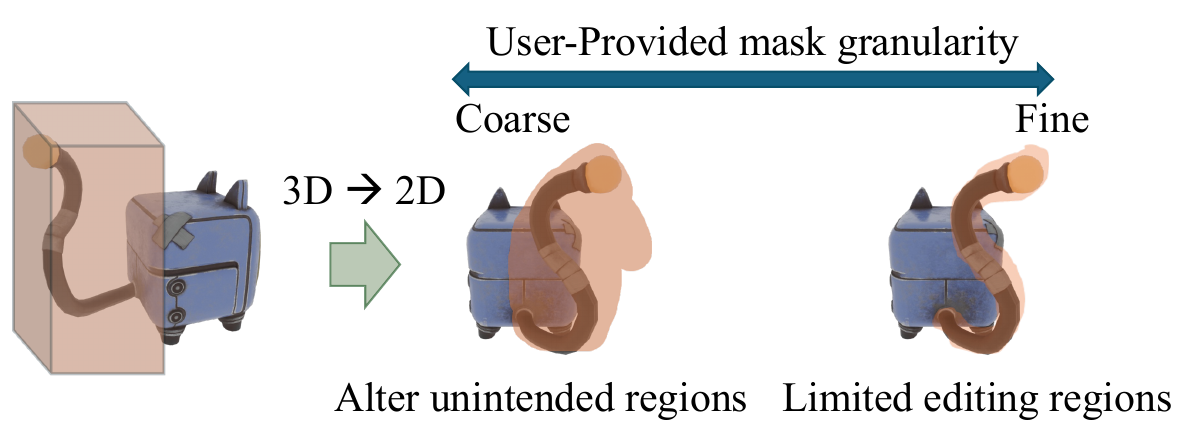}
    \vspace{-8mm}
    \caption{
    \textbf{Ambiguous intended regions.}
    The intended region to be edited is clear in 3D (e.g. the cat tail). However, after projecting to 2D, regardless of the granularity of the user-provided masks, the editing will either alter some unintended regions (e.g. the robot cat) or be too limited for reasonable editing.
    }
    \label{fig:occlusion}
    \vspace{-5mm}
\end{figure}

%% file: figure_tex/overview.tex
\begin{figure*}
  \centering
  \includegraphics[width=\linewidth]{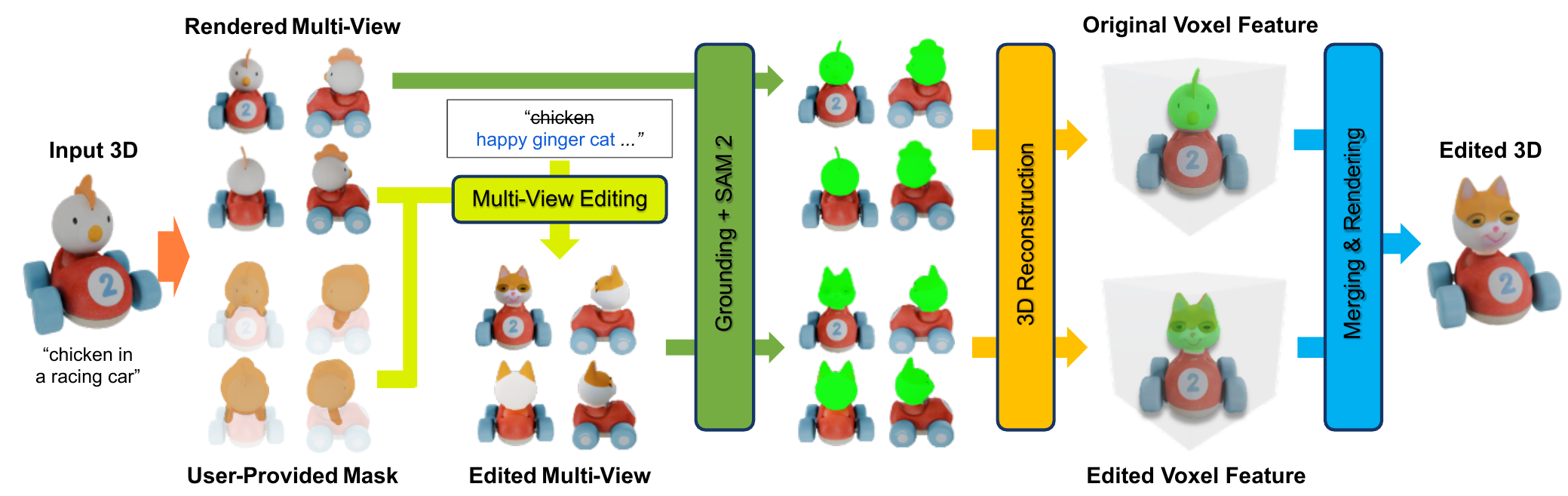}
  \vspace{-4mm}
  \caption{\textbf{Overview of \OURS{}}. 
  Given an input 3D object, we first render its multi-view images from 4 orthogonal views. We then obtain editing input from the user, describing in text as well as rough 2D masks the desired edits.
  We perform synchronized multi-view editing based on the text prompts as well as the user-provided masks (Sec.\ref{sec:3.1}.
  Due to the rough masks and the unclear intended regions caused by ambiguous 3D-2D projection, we detect the intended regions with Grounding Dino and SAM 2 (Sec.\ref{sec:3.2}, where the segmentation results are lifted to 3D for the final merging operation (Sec.\ref{sec:3.3}).
  }
  \label{fig:overview}
  \vspace{-2mm}
\end{figure*}

%% file: sec/2_related_hy.tex
\section{Related Works}
\label{sec:related}

\smallskip\noindent\textbf{2D Editing} applies global or local modifications to an image based on user instructions. It has gained significant attention for enabling a more interactive user experience in content creation. 
To achieve this, existing methods either fine-tune text-to-image models with specialized instructional editing datasets~\cite{instructp2p,magicbrush}, or use training-free approaches with inpainting~\cite{blendeddiffusion,glide} or cross-attention mechanisms~\cite{masactrl,prompt2prompt,attendandexcite}.
To edit user-provided images, various diffusion inversion techniques have been developed. 
For example, SDEdit~\cite{sdedit} introduces noise to images to capture intermediate steps in the diffusion process, null-text inversion~\cite{nulltext} inverts the deterministic DDIM inversion~\cite{ddim}, and recent works~\cite{ledits++,editfriendlyddpm} use DDPM inversion~\cite{ddpm} to enhance editing capabilities.

In this work, one step to achieve 3D editing is to perform 2D multi-view image editing. We apply
the DDPM inversion~\cite{ddpm} and Prompt2Prompt~\cite{prompt2prompt} to operate within a diffusion-based sparse multi-view generation model.

\smallskip\noindent\textbf{3D Reconstruction from Sparse Multi-Views} refers to the task of reconstructing a 3D instance from a limited number of multi-view images. To achieve this, Score Distillation Sampling (SDS)~\cite{dreamfusion} and its variants~\cite{prolificdreamer,sjc,fantasia3d,magic3d,magic123} optimize a 3D scene representation by reconstructing the given sparse views and generating novel views through gradients from large-scale pre-trained text-to-image diffusion models~\cite{ldm}. However, these approaches are often computationally demanding.
Beyond per-scene optimization, PixelNeRF \cite{pixelnerf} generates a NeRF representation \cite{nerf} of a scene from sparse-view images by training the model across multiple scenes. With recent advancements in large-scale 3D datasets such as Objaverse \cite{objaverse}, follow-up works \cite{wonder3d, lrm, instant3d, dmv3d, crm, lgm, gtr,upfusion} have improved this feed-forward 3D reconstruction approach for object-centric 3D assets. In this work, we utilize one recent method, GTR \cite{gtr}, to quickly generate 3D meshes from 4 views.

\smallskip\noindent\textbf{3D Editing} presents additional challenges compared to 2D editing due to the need to maintain spatial consistency in 3D. To address this, some methods train 3D editing models using paired 3D datasets \cite{shapetalk, instructp2p}, however, these approaches are limited by a lack of diverse and complex datasets. Other methods adapt 2D editing models, such as InstructPix2Pix \cite{instructpix2pix}, for the 3D domain \cite{instructnerf2nerf, tailor3d}, by iteratively updating multi-view images of a scene. However, without synchronized multi-view updates, this approach often results in flickering or inconsistent views.
Alternatively, some methods propose a generation-reconstruction loop to modify 3D representations using intermediate denoised images \cite{mvedit} or SDS gradients \cite{voxe, shapeditor, interactive3d, paintbrush} in the diffusion process. While these methods can achieve 3D consistency, they often struggle with quality or suffer from high computational costs.
In our work, we perform multi-view editing and reconstruct the edited object in 3D. Beyond that, to preserve unchanged regions, we carefully design an approach by detecting the edited 3D regions and integrating the intended 3D edited regions into the original shapes.

%% file: sec/3_method.tex
\section{Method}
\label{sec:method}
We aim to achieve fast 3D asset editing in a training-free manner, allowing for precise and user-guided edits. 
Our approach achieves this through multi-view image editing in 2D, followed by lifting the 2D edits into 3D. This process can be summarized into 3 main steps: (1) synchronized sparse multi-view editing in 2D (Sec.\ref{sec:3.1}), (2) detecting intended editing regions across 2D views through the Grounding Dino \cite{groundingdino} and SAM 2 \cite{sam} approach (Sec.\ref{sec:3.2}), and (3) lifting the intended editing regions to 3D and merging the edited shape into the original (Sec.\ref{sec:3.3}). 
Our approach is illustrated in Fig.~\ref{fig:overview}.

\subsection{Synchronized Sparse Multi-View Editing}
\label{sec:3.1}
To edit a given 3D object $\mathcal{O}$, we leverage the power of 2D editing through multi-view diffusion models. 
We first perform synchronized sparse multi-view edits using a pre-trained multi-view diffusion model.
In practice, we use MVDream \cite{mvdream}, which generates 4 orthogonal views based on a text prompt. Note that our approach remains agnostic to the specific diffusion-based multi-view generation model used.
 
We first render multi-view images from $\mathcal{O}$, then apply the DDPM diffusion inversion mechanism \cite{editfriendlyddpm} to revert the multi-view images to their initial noise vectors, denoted as $\bm{x^T}$, where $T$ is the number of diffusion timesteps during the denoising process. We will use these vectors $\bm{x^T}$ as the initial latent vectors for the editing process in diffusion.
We denote the text prompt for the input shape and the edited shape as $y_i$ and $y_e$, respectively. 
Also, to enable better editing control and interaction, we further take user-provided masks in 4 views as input. These masks indicate target regions for the edit, denoted as $\bm{M_\text{user}} \in \mathbb{R}^{4\times H \times W}$ where $H$ and $W$ are the height and weight of the images. Note that our method does not require precise and accurate masks.

To edit multi-view images, we apply the Prompt-to-Prompt \cite{prompt2prompt} approach on the multi-view diffusion model. 
 For simplicity, we present the basic operation at each diffusion timestep and within each attention block. To be specific, Prompt-to-Prompt~\cite{prompt2prompt} generates an edited latent vector $\bm{x_e}'$ by replacing the self- and cross-attention weights of the original latent vector, $\bm{x_i}$, with the edited prompt $y_e$. To confine modifications to the desired regions, we blend the latent vectors $\bm{x_e}'$ and  $\bm{x_i}$ using user-provided masks $\bm{M_\text{user}}$. 
We denote the final edited latent vector at each step as $\bm{x_e}$:
\begin{equation}
    \bm{x_e} \leftarrow \bm{M_\text{user}} \cdot \bm{x_e}' + (1- \bm{M_\text{user}}) \cdot \bm{x_i}.
    \label{eq:1}
\end{equation}
In practice, we downsample the user-provided masks $\bm{M_\text{user}}$ to match the feature resolution at each model layer, and the edited latent vector, $\bm{x_e}$, serves as the input to the next model layer.

Using the inverted noise vector $\bm{x^T}$ ensures that the edited results align with the original texture style. 
However, the masks $\bm{M_\text{user}}$ are often imprecise or misaligned with the target regions, which can still affect regions that are not intended to be altered. Furthermore, occlusions along the depth dimension introduce additional challenges in accurately and semantically localizing the intended editing regions in 3D. We illustrate the issue in Fig.~\ref{fig:occlusion}.
To address this, we apply an automated grounding approach that
detects the intended editing region in both 2D and 3D.

\subsection{Detection of Intended Editing Regions in 2D}
\label{sec:3.2}

The intended editing regions refer to the specific semantic areas that correspond to the editing prompt. For example, in Fig.~\ref{fig:teaser}, changing ``chicken" to ``cat" implies that the desired editing region pertains only to the areas representing the ``chicken" and ``cat" concepts. Regions (both 2D and 3D) that are not semantically related to these concepts should remain unchanged.

To ensure that only intended editing regions are edited while allowing rough user-provided masks, we propose to detect the intended editing regions by applying Grounding Dino~\cite{groundingdino} and SAM 2~\cite{sam} to both the original and the edited multi-view images. 
To begin with, we identify the changing concept by comparing the original prompt $y_i$ and the editing prompt $y_e$. We then localize the changing concept within the user-provided mask regions $\bm{M_\text{user}}$ and obtain corresponding bounding boxes in multiple views. Formally, we write the procedure as
\begin{equation}
    \bm{\text{bbox}_x} \leftarrow \text{Grounding}(\bm x, \bm{M_\text{user}}, y_i, y_e),
\end{equation}
where $\bm{\text{bbox}_x}$ are the bounding boxes for the changing concept in the multi-view images $\bm x \in \{\bm{x_i}, \bm{x_e}\}$.\footnote{We use $\bm x$ to represent both latent vectors and images, without distinguishing between the two.}

Afterward, we segment and track the changing elements across views using the grounding bounding boxes $\bm{\text{bbox}_x}$. Formally, this procedure can be described as $\text{SAM}(\bm x, \text{bbox}_x)$. This process yields the intended editing regions in segmentation format for both the original and edited multi-views.  

\input{alg_tex/alg_merge}
\input{figure_tex/exp_compare}

\subsection{Lifting and Merging Edits in 3D}
\label{sec:3.3}

Finally, we lift the 2D edits to 3D and merge the 3D-edited regions into the original shape.

\smallskip\noindent\textbf{3D Segmentation by Lifting 2D Segmentations}. We mark the intended editing regions 
on multi-view images using a green color.
The color-coded multi-view images are then reconstructed in 3D using an offline 3D reconstruction model~\cite{gtr}. This model takes multiple views as input and represents the shape as a triplane feature. 
Two separate decoders—one for Signed Distance Function (SDF) and one for color— generate a geometry field and a color field, respectively. 
Through this process, we create a \textit{3D segmentation field} from the color-coded multi-view images.
For each 3D position within the 3D space represented by the triplane, we determine whether the position lies within the 3D intended editing regions based on its color value. That is, we apply a decision threshold for the distance between each color value and the preset green color to identify targeted regions. This produces two 3D masks, indicating the intended editing regions in 3D for both the original shape and the edited shape, denoted as $\bm{M_i}$ and $\bm{M_e}$, respectively.

\input{figure_tex/exp_ourresult}

\smallskip\noindent\textbf{Merging Edits in 3D}. 
As aforementioned, coarse user-provided masks and occlusion issues in 2D projections often result in unwanted alterations, compromising the preservation of unaffected 3D regions.
Therefore, the directly reconstructed shapes from the edited multi-view images using the reconstruction model~\cite{gtr} cannot be the final editing output, as illustrated in Fig.~\ref{fig:comp_merge}. 

Using the 3D reconstruction model~\cite{gtr}, we extract voxel features for both original and edited shapes by interpolating their triplane features, denoted as $\bm{V_i}$ and $\bm{V_e}$ for the original and edited shapes, respectively. Both $\bm{V_i}, \bm{V_e} \in \mathbb{R}^{A \times A \times A \times F}$, where $A$ is the voxel resolution, and $F$ is the feature dimension, $A = 256$ and $F = 40$ in practice.

To merge 3D features, we first nullify the original specific regions $\bm{M_i}$ from the original voxel feature $\bm{V_i}$, and then replace the target edited regions $\bm{M_e}$ with edited feature $\bm{V_e}[\bm{M_e}]$. We write the above operations as follows:
\begin{equation}
    \bm{V_i}[\bm{M_i}] \leftarrow  \emptyset , \text{ and } 
    \bm{V_i}[\bm{M_e}] \leftarrow  \bm{V_e}[\bm{M_e}].
\end{equation}

We refer to this approach as a naive copy-paste method. While theoretically plausible, we observe that this straightforward approach typically introduces discontinuities at
the 3D editing boundaries, as shown in Fig.~\ref{fig:comp_merge}. To address this, we propose an averaged merging approach that provides a more robust blend of 3D features. In the improved method, we dilate the 3D mask $\bm{M_e}$ by a dilation $d$ and then use an exclusive or (a.k.a. $\text{XOR}$) operation to select the boundary mask regions for smooth blending. Next, we linearly interpolate the two voxel features $\bm{V_i}$ and $\bm{V_e}$ within the boundary regions using a coefficient $\theta$, in practice $\theta=0.5$. We illustrate the merging process in Alg.~\ref{alg:merge}. After merging, we generate a textured mesh from the blended voxel feature, using the decoders in the 3D reconstruction model~\cite{gtr}.

%% file: alg_tex/alg_merge.tex
\begin{algorithm}[t]
\DontPrintSemicolon
\caption{Merging Voxel Features}\label{alg:merge}

\KwInput{$\bm{V_i} \text{ and } \bm{V_e} \in \mathbb{R}^{A \times A \times A \times F}, \bm{M_i} \text{ and } \bm{M_e} \in \mathbb{R}^{A \times A \times A}, d \in \mathbb{N}, \text{ and } 
\theta \in [0, 1]$ }

\KwOutput{$\bm{V_\text{blend}} \in \mathbb{R} ^{A \times A \times A \times F}$}  
    $\bm{V_i}[\bm{M_i}] \leftarrow \emptyset$  \label{alg:merge:cps} \\  
    $\bm{V_i}[\bm{M_e}] \leftarrow  \bm{V_e}[\bm{M_e}]$  \label{alg:merge:cpe} \\
    $\bm N \leftarrow  \text{Dilation}(\bm{M_e}, d)$  \\  \label{alg:merge:blends}
    $\bm K \leftarrow  \bm{M_e} \oplus \bm N $  \tcp{$\oplus$ means XOR}
    $\bm{V_\text{blend}} \leftarrow \bm{V_i}$ \\
    $\bm{V_\text{blend}}[\bm K] \leftarrow  \theta \odot \bm{V_i}[\bm K] + (1 - \theta) \odot \bm{V_e}[\bm K]$  \\
    \textbf{return} $\bm{V_\text{blend}}$
\end{algorithm}

%% file: figure_tex/exp_compare.tex
\begin{figure*}
    \centering
    \begin{minipage}{0.12\linewidth}
    \centering
        \vspace{.7cm}
        Input
        \vspace{2.2cm}
        
        Tailor3D~\cite{tailor3d}
        \vspace{2.2cm}
        
        MVEdit~\cite{mvedit}
        \vspace{2.2cm}
        
        Vox-E~\cite{voxe}
        \vspace{2.0cm}
        
        Ours
    \end{minipage}
    \begin{minipage}{0.87\linewidth}
    \includegraphics[width=\linewidth]{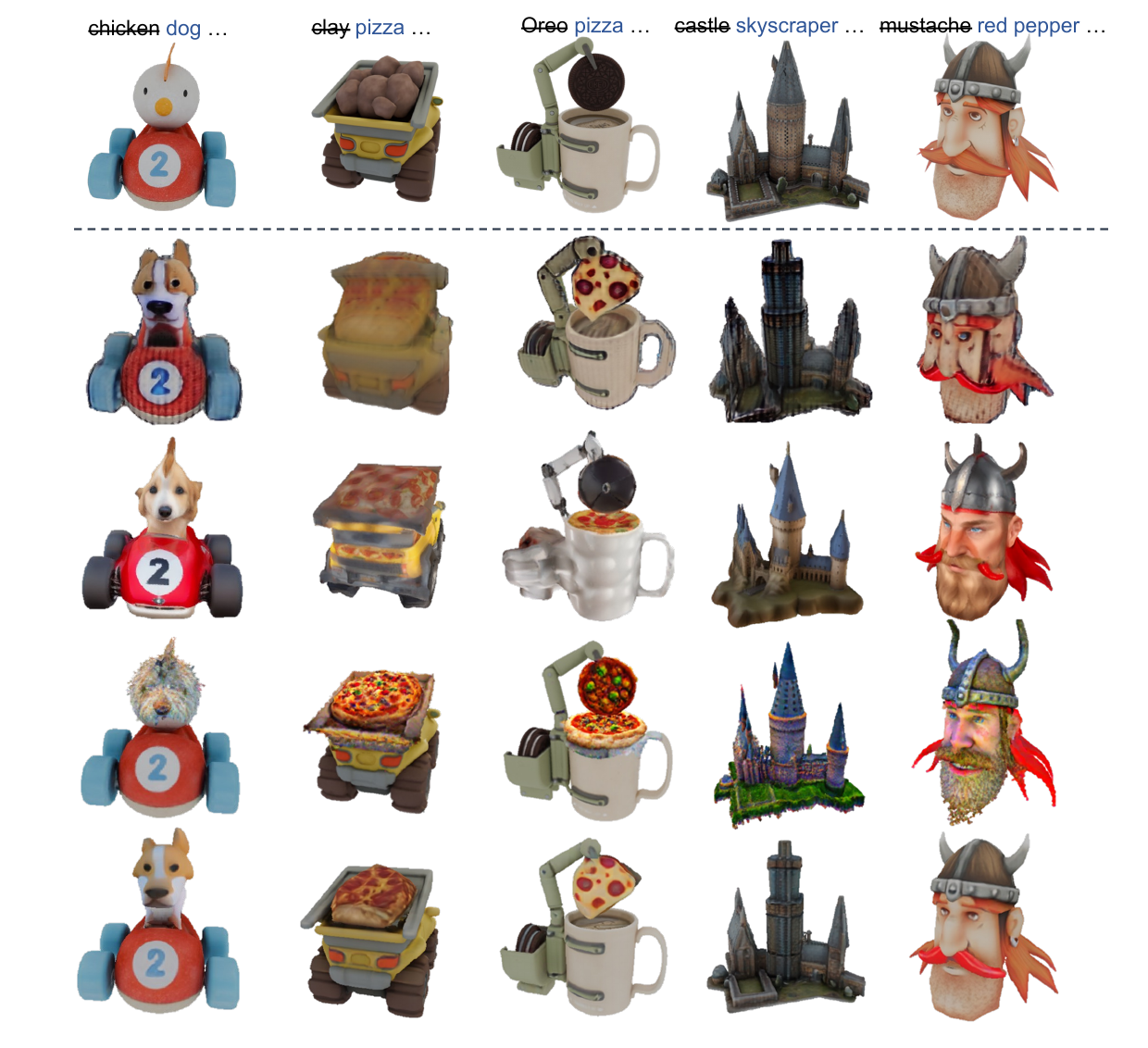}
    \end{minipage}
    \vspace{-0.4cm}
    \caption{\textbf{Qualitative comparison}. Our method can perform diverse editing samples and only edit the intended regions.  }
    \label{fig:qual_comp}
\end{figure*}

%% file: figure_tex/exp_ourresult.tex
\begin{figure*}
    \centering
    \includegraphics[width=0.95\linewidth]{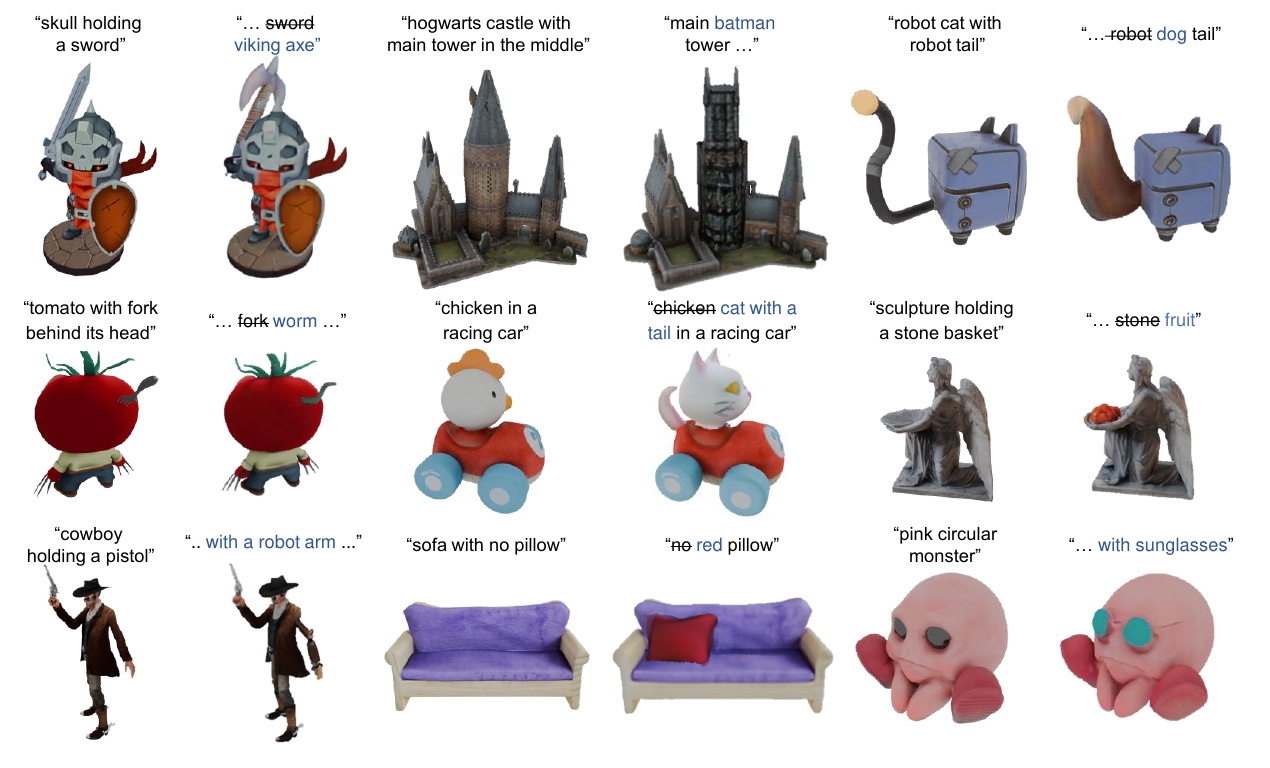}
    \vspace{-6mm}
    \caption{\textbf{More editing results from \OURS{}}. Our method can perform a wide range of editing on various shapes.
    }
    \vspace{-2mm}
    \label{fig:our_results}
\end{figure*}

%% file: sec/4_results.tex
\input{table_tex/table_quan}
\section{Experiments}
\label{sec:experiments}

\subsection{Evaluation} \label{sec:eval}
Our evaluation dataset contains 18 unique shapes and 40 editing prompts. We use shapes from GSO \cite{gso} and Objaverse~\cite{objaverse}. We evaluate our method based on the quality of the editing and consistency with the input shapes.

We use the GPTEval3D~\cite{gpteval3d} metric to evaluate the quality of edited shapes and their alignment with the text prompts. GPT-4V is provided with multi-view renderings of two methods at a time, and instructed to pick one based on text-prompt alignment, 3D plausibility, and texture details. There were 120 total questions, each answered 3 times.

Since the GPTEval3D metric does not consider the input shape, it cannot measure whether the shape remained intact, and whether the edited shape is consistent with the high-level style of the input shape. Therefore, we adopt the directional CLIP~\cite{clip} score metric, $\text{CLIP}_\text{dir}$ from previous works~\cite{gal2022stylegan, voxe}. 
$\text{CLIP}_\text{dir}$ evaluates the average difference between text feature change direction and image feature change direction, where images are multi-view renderings of input and edited shapes. To ensure our evaluation is not affected by a particular implementation of this metric, we introduce three variants. $\text{CLIP}_\text{dir-cos}$ replaces the text-image direction vector difference with cosine distance, while $\text{CLIP}_\text{dir-avg}$ and $\text{CLIP}_\text{dir-avg-cos}$ compute the same metrics by averaging image vectors first rather than scores.

We finally introduce two additional directional metrics. $\text{CLIP}_\text{diff-edit}$ is the CLIP score difference between input and output image-text pairs concerning only the edited part of the input and output text prompts. $\text{CLIP}_\text{diff-noedit}$ is the CLIP score difference between input and output using a fixed text prompt where the edited part of the input text is replaced with a generic word, i.e., ``object."
These metrics enforce that the CLIP text matching scores are preserved between the input and edited shapes, both for edited and unedited parts of the text.

We report  all directional metrics multiplied by 100 for  higher precision.
We refer readers to our supplementary material for further details about the evaluation metrics. 

\input{figure_tex/exp_diverse}
\subsection{Results} \label{sec:results}
Our method can generate various impressive edited shapes from complex input shapes and prompts. We illustrate a variety of our results in Fig.~\ref{fig:our_results}. 
Our approach flexibly edits various different elements of the 3D objects, for instance replacing a ``sword" of a skull warrior with a ``viking axe," resulting in coherent, seamless edits in both texture and geometry. 
Our edits also follow the structure of the input shape when applicable; for instance, when replacing a curvy fork with a worm, the worm maintains the same curved structure as the initial fork.
We can also insert new objects, such as a ``pillow" or ``sunglasses." Our method even enables both replacement and addition at the same time as in ``cat with a tail" example, replacing the chicken with a cat and simultaneously placing a tail at the back of the car.

Given the same prompt, our method can generate different results with different seeds. In Fig.~\ref{fig:multiple_gen}, we show generations of ``cat" and ``dog" samples with various seeds. The resulting shapes vary in their head, eye, and ear structures with different colors and sizes. We can further control the various aspects of the generated shape through user prompts. This is also shown in Fig.~\ref{fig:multiple_gen}, where we adjust the mood and appearance of the generated shape.

\smallskip\noindent\textbf{Comparison with Baselines.}
Fig.~\ref{fig:qual_comp} shows a qualitative comparison of our method against several state-of-the-art methods: Tailor3D~\cite{tailor3d}, MVEdit~\cite{mvedit} and Vox-E~\cite{voxe}. 
Our approach shows significant improvements, in both editing quality as well as consistency with the original shape. 
 
Similar to our method, Vox-E allows controllable editing through merging but at the expense of an expensive SDS-based optimization that can tend towards more global changes than local ones.  Since Tailor3D accepts edited front and back views as input, we ran their method using our multi-view editing results. Tab.~\ref{tab:quant_comp} shows a comparison using GPTEval3D. 
While our method is consistently preferred, improvements are not as large since this metric does not measure consistency with the input. A method could globally change the shape and still achieve better results. This is because this metric does not take input shape into account and only considers edited output and the prompt. To complement this metric, we calculated the directional CLIP score and its few variants in Tab.~\ref{tab:clip_dir}; this considers consistency with the input, and demonstrates that our approach achieves significant improvements over the baselines.

\smallskip\noindent\textbf{Perceptual Study.} We prepare a perceptual study to compare our method with three other baselines asking users three different questions: ``\textit{Select the one that follows the following prompt more closely}", ``\textit{Select the one with better visual quality}",  ``\textit{Which example better preserves the parts that were not instructed to be edited with the prompt?}". There are 360 questions in total, each answered by 10 different participants, totaling 3600 responses. Results are presented in Tab.~\ref{tab:user_study}. In all questions, our method is preferred over the baselines.

\smallskip\noindent\textbf{Runtime Analysis.} Our method enables fast iteration, taking around 24 seconds to obtain initial multi-view editing results. Merging then takes another 50 seconds to produce a final refined shape. 
Tab.~\ref{tab:runtime} shows a comparison with baselines, using a single RTX 3090 for measurements, except for MVDream, which we run on RTX A6000. 
Tailor3D~\cite{tailor3d}, concurrent to our work, also operates fast, taking 2 seconds for a forward pass using our multi-view editing results as input (26 seconds in total). MVEdit~\cite{mvedit} does not employ any merging, performing editing in around 6 minutes. Since Vox-E~\cite{voxe} involves a long SDS optimization process, its overall inference can take around an hour.

\input{figure_tex/exp_merging}

\input{table_tex/table_ablation}

\subsection{Ablations}
Tab.~\ref{tab:quant_ablation} and Fig.~\ref{fig:comp_merge} ablate our merging approach, measuring the chamfer distance between the edited and input shapes.

\smallskip\noindent\textbf{Shape Preservation through Merging.}
Our merging algorithm ensures that only the regions described by the user through a mask and prompt changes. In this ablation study, we only do multi-view editing, and leave out the merging operation. As shown in Fig.~\ref{fig:comp_merge}, without any merging, regions that are not intended by the user can change. In the ``chicken in a racing car" example, when the user replaces the chicken with a cat, some part of the car is also altered since the user mask covers that area. In our merging step, we detect the changed region (``cat") and erased region (``chicken") so that we keep the rest of the shape (``car") fixed.

\smallskip\noindent\textbf{Average Merging.}
After edited regions are detected, we merge the voxel grids of the input and edited reconstructions to preserve consistency with the input. In contrast, Vox-E \cite{voxe} uses copy-pasting for merging. That is, they copy the detected part from the edited shape and paste it into the original shape. However, a simple copy-paste approach can create boundary artifacts such as gaps between the edited region and the original shape, as shown in Fig.~\ref{fig:comp_merge}. To fix these boundary problems, we dilate the masks. Within the dilated region, we take the average of the edited shape and the original shape, which provides a smoother transition.

\smallskip\noindent\textbf{Limitations.}
Although our method can generate high-quality editing results, we are limited by the 256x256 resolution of the multi-view diffusion model, MVDream~\cite{mvdream}. In addition, our method currently focuses on 3D assets that can be rendered from four inward-facing views. However, this assumption cannot effectively capture large-scale scenes, such as indoor rooms where more views within the scene are needed.

%% file: table_tex/table_quan.tex
\begin{table}[t]
\centering
\normalsize
\resizebox{0.49\textwidth}{!}{
\begin{tabular}{ccccc}
\hline
Method & Prompt Algn. & 3D Plausibility & Texture & Overall \\ \hline
Tailor3D~\cite{tailor3d} & 98$\%$ & 99$\%$ & 99$\%$ & 99$\%$ \\
MVEdit~\cite{mvedit} & 57$\%$ & 55$\%$ & 55$\%$ & 57$\%$ \\
Vox-E~\cite{voxe} & 53$\%$ & 68$\%$ & 50$\%$ & 55$\%$ \\
\hline
\end{tabular}}
\caption{\textbf{Comparison using GPTEval3D~\cite{gpteval3d}}. Scores indicate the percentage of our method being selected over baselines.}
\label{tab:quant_comp}
\vspace{-2mm}
\end{table}

\begin{table}[t]
\centering
\normalsize
\resizebox{0.49\textwidth}{!}{ 
\begin{tabular}{ccccc}

\hline
 & Prompt Algn. & Visual Quality & Preserving Shape \\ \hline
Tailor3D~\cite{tailor3d} & 96$\%$ & 97$\%$ & 99$\%$ \\
MVEdit~\cite{mvedit} & 68$\%$ & 68$\%$ & 97$\%$ \\
Vox-E~\cite{voxe} & 78$\%$ & 88$\%$ & 89$\%$ \\
\hline
\end{tabular}}
\caption{\textbf{User study results comparing our method against baselines.} The percentage shows the preference for our method.}
\label{tab:user_study}
\vspace{-2mm}
\end{table}

\begin{table*}[t]
\centering
\normalsize
\resizebox{0.95\textwidth}{!}{
\begin{tabular}{ccccccc}
\hline
Method & $\text{CLIP}_\text{dir}\uparrow$ & $\text{CLIP}_\text{dir-cos}\uparrow$ & $\text{CLIP}_\text{dir-avg}\uparrow$ & $\text{CLIP}_\text{dir-avg-cos}\uparrow$ & $\text{CLIP}_\text{diff-edit}\downarrow$ & $\text{CLIP}_\text{diff-noedit}\downarrow$
\\ \hline
Tailor3D~\cite{tailor3d} & 1.416 & 3.710 & 1.417 & 4.593 & 12.050 & 5.619 \\
MVEdit~\cite{mvedit} & 0.782 & 2.578 & 0.783 & 3.164 & 10.014 & 4.118 \\
Vox-E~\cite{voxe} & 1.622 & 6.178 & 1.621 & 8.153 & 10.528 & 3.432 \\
\OURS{} (Ours) & \textbf{1.782} & \textbf{7.679} & \textbf{1.782} & \textbf{11.554} & \textbf{8.812} & \textbf{2.636} \\
\hline
\end{tabular}}
\vspace{-2mm}
\caption{\textbf{Directional CLIP score metrics~\cite{voxe} for evaluating editing fidelity and prompt consistency}. Our method outperforms baselines across all directional CLIP metrics. Metrics are scaled by 100 to ease reading and allow for more precision. }
\vspace{-3mm}
\label{tab:clip_dir}
\end{table*}

\begin{table}[t]
\centering
\begin{tabular}{cccc}
\hline
Method & Editing & Merging & Total \\ \hline
Tailor3D~\cite{tailor3d} & 26 sec & - & 26 sec \\
MVEdit~\cite{mvedit} & 6 min & - & 6 min \\
Vox-E\cite{voxe} & 60 min & 15 min & 75 min \\
\OURS{} (Ours) & 24 sec & 50 sec & 74 sec \\
\hline
\end{tabular}
\caption{\textbf{Runtime comparison.} We measure the runtime of our baseline methods.}
\vspace{-4mm}
\label{tab:runtime}
\end{table}

%% file: figure_tex/exp_diverse.tex
\begin{figure}
    \centering
    \includegraphics[width=0.86\linewidth]{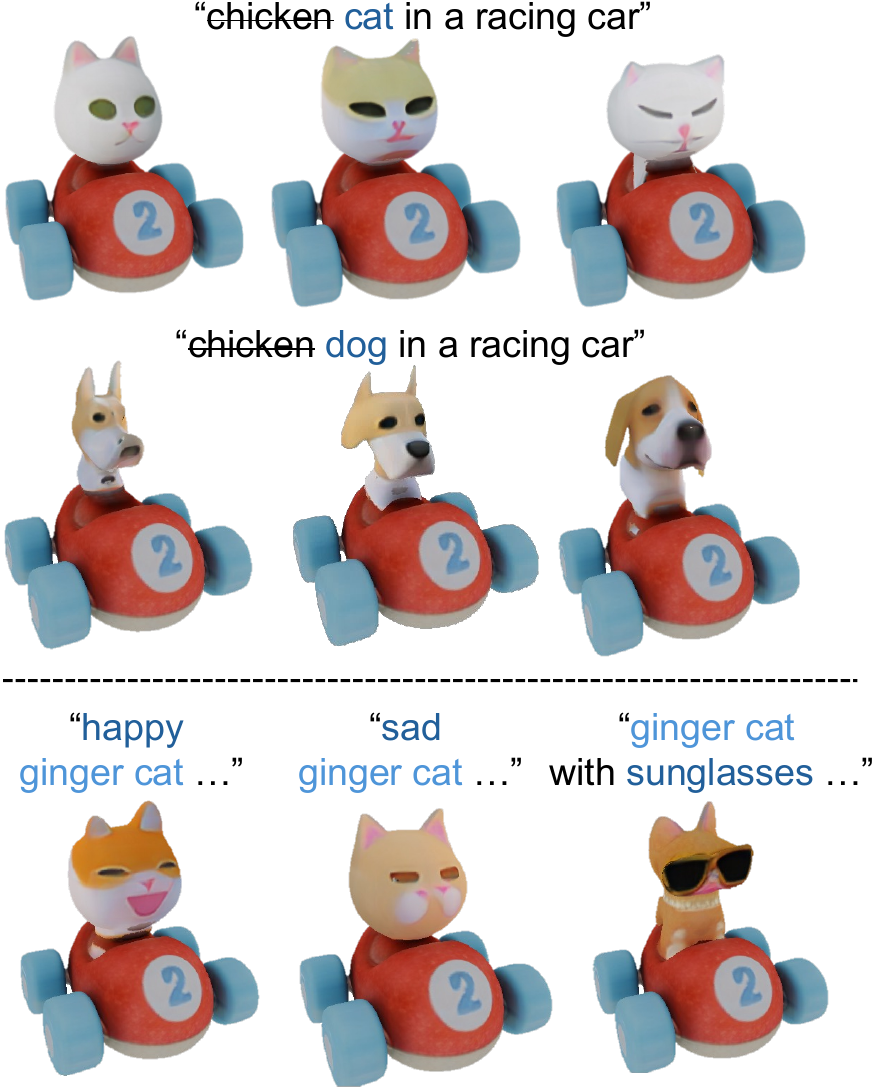}
    \caption{\textbf{Multiple generations and detailed control through prompt}. Our method can generate different results for the same prompt using different seeds. Moreover,  Our method can handle detailed prompts that can modify various aspects of the shape such as appearance and mood. For instance, here we can define the type of the cat (e.g. ginger cat) and the mood (e.g., happy).}
    \label{fig:multiple_gen}
    \vspace{-2mm}
\end{figure}

%% file: figure_tex/exp_merging.tex
\begin{figure}
    \centering
    \includegraphics[width=\linewidth]{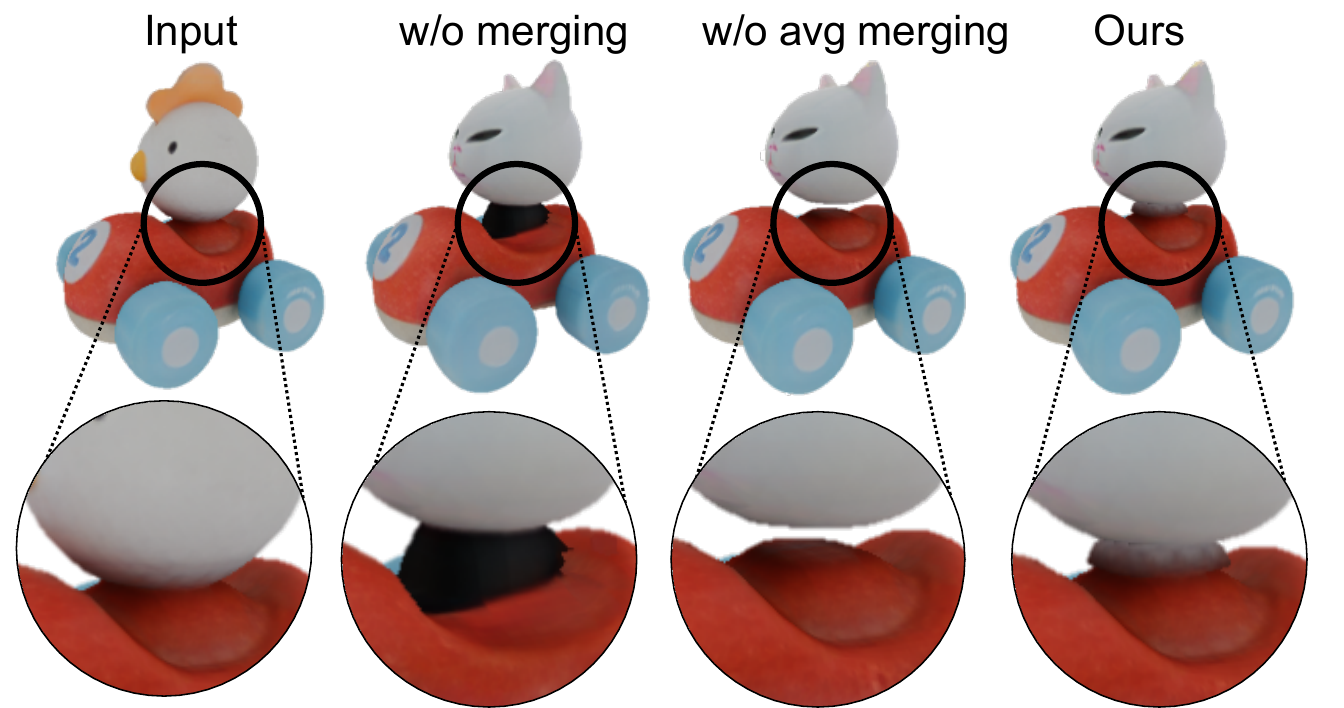}
    \caption{\textbf{Qualitative ablation of our merging algorithm}. We can keep the original parts of the input fixed. Here when we insert a cat, the editing breaks the neighboring regions. Thanks to our merging algorithm, we can recover the original parts of the shape.  }
    \label{fig:comp_merge}
    \vspace{-2mm}
\end{figure}

%% file: table_tex/table_ablation.tex
\begin{table}
\centering
\begin{tabular}{cc}
\toprule
Method & Chamfer Distance $\downarrow$ \\
\midrule
w/o Merging  & 2.95\\
w/o Average Merging  & 2.29 \\
Ours &  \textbf{2.28} \\
\bottomrule
\end{tabular}
\vspace{-2mm}
\caption{\textbf{Quantitative ablation study of our merging algorithm}. We calculate the chamfer distance to the input shape for each ablation. Chamfer Distance value is multiplied by $10^3$. Our algorithm is effective in keeping the edited shape consistent with the input. }
\vspace{-3mm}
\label{tab:quant_ablation}
\end{table}

%% file: sec/5_conclusion.tex
\section{Conclusion}
\label{sec:conclusion}
We propose a fast and controllable 3D editing method that can handle a wide variety of 3D shapes and editing prompts. We employ the editing strength of powerful multi-view models, lift edits to 3D, and merge edits in 3D in order to ensure unedited regions remain consistent with the input shape. 
Hence, our method produces high-quality editing results with fast runtime speeds.
We believe this shows a significant potential for high-quality, controllable, seamless, and fast 3D editing.
\\
\\

\noindent \textbf{Acknowledgments} This work is partially done during Ziya's and Can's internships at Snap. Matthias Nie{\ss}ner was supported by the ERC Starting Grant Scan2CAD (804724) and Angela Dai was supported by the ERC Starting Grant SpatialSem
(101076253).

%% file: sec/X_suppl.tex
\clearpage
\setcounter{page}{1}
\section{Appendix}
We present additional details about \OURS{} in this appendix. We start by explaining some of the implementation details in Sec.~\ref{sec:supp_impl_details}. In Sec.~\ref{sec:supp_auto_mask}, we discuss automatic masking, an alternative to user-brushed masking. Sec.~\ref{sec:supp_mask_gran} follows this discussion with the effect of mask granularity on the editing process. Finally, we explain the directional CLIP metrics we used for baseline comparison in Sec.~\ref{sec:supp_eval_metrics}.

\subsection{Implementation Details}
\label{sec:supp_impl_details}
We used the official implementation and checkpoint of MVDream as our multi-view diffusion model. It has 256 x 256 resolution and it can generate four views by default. In all of our generations, we set the classifier-free guidance scale of the diffusion process to 10. Official DDPM inversion~\cite{editfriendlyddpm} implementation only handles single-image but we modified it to handle our four view renderings. The inversion process takes 9 seconds on RTX 3090. With the inverted latents, we ran our inference for 41 steps, which takes around 12 seconds on an RTX 3090. For the segmentation, we calculate bounding boxes using Grounding DINO~\cite{groundingdino} for all views and add these as constraints to SAM 2~\cite{sam} tracking. That is to help SAM 2 with the segmentation, we constrain each frame separately. For merging and reconstruction, we modify GTR~\cite{gtr}, which is a feed-forward reconstruction model. GTR mainly operates on triplanes but just before reconstruction, those triplanes are converted into a voxel grid. We manipulated the voxel grid it generated to merge two different shapes.

\subsection{Automatic Masking} 
\label{sec:supp_auto_mask}

In addition to user-brushed masks, we can also generate and operate on automatically generated masks. Even though they limit the editing region, when compared to user-brushed masks; they can be practically used as a starting point for user-brushed masking. 

\begin{figure}
    \centering
    Automatically Generated Mask \hspace{.4cm}  User-Brushed Mask
    \includegraphics[width=\linewidth]{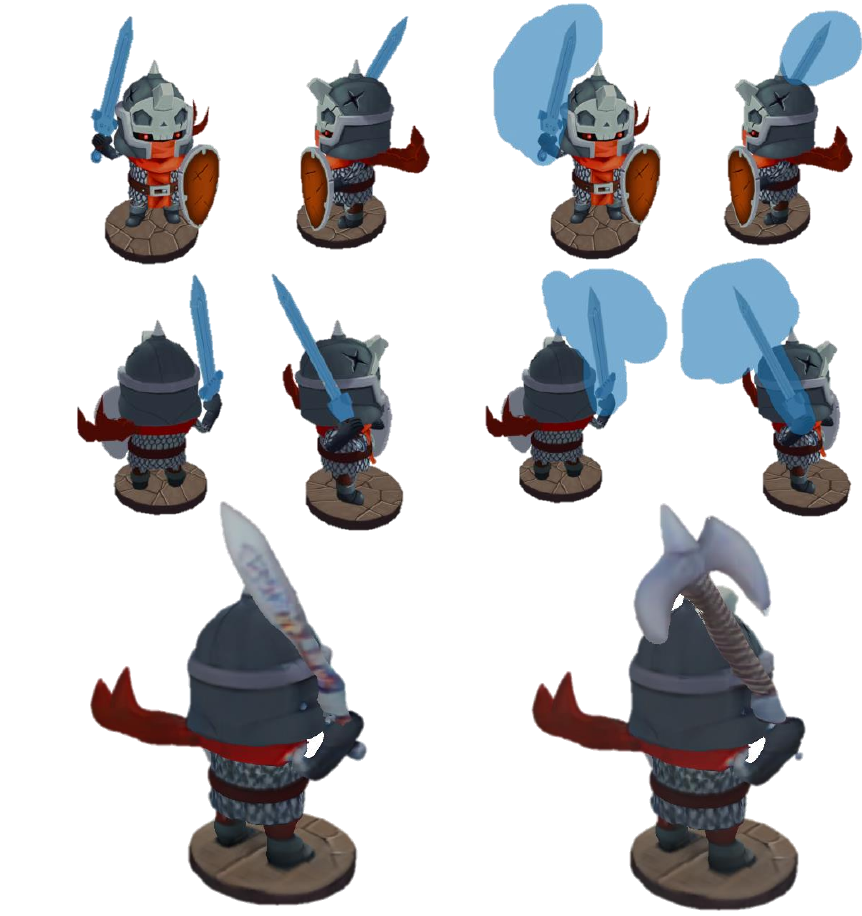}
    \caption{\textbf{Comparing automatically generated mask to user-generated mask}. Users may want to do specific editing such as replacing the \textit{``sword"} with \textit{``a viking axe"}. If we only rely on automatic masking, the result may not follow the user's intention since the automatically generated mask can limit the editing to a certain region. However, when we rely on explicit masking, we can get the specific shape requested by the user.}
    \label{fig:auto_masking}
\end{figure}

We leverage our segmentation approach to replace masks given by the user. We use an input prompt from the user to detect the target region using Grounding DINO~\cite{groundingdino} and SAM 2~\cite{sam}. This segmentation method gives us a mask restricted only to the sword. As a result, the generation process cannot go beyond that region. However, when we accept input from user masks, user can explicitly show their intention with the mask and can generate a \textit{"viking axe"}, as shown in  Fig.~\ref{fig:auto_masking}.

We want to reiterate that although the user-brushed masks are too coarse and not 3D-consistent, our method can generate impressive results without modifying the original parts of the shape. That is, a quickly drawn mask is enough for our method to work.

\subsection{Mask Granularity}
 \label{sec:supp_mask_gran}

\begin{figure}
    \centering
    \includegraphics[width=\linewidth]{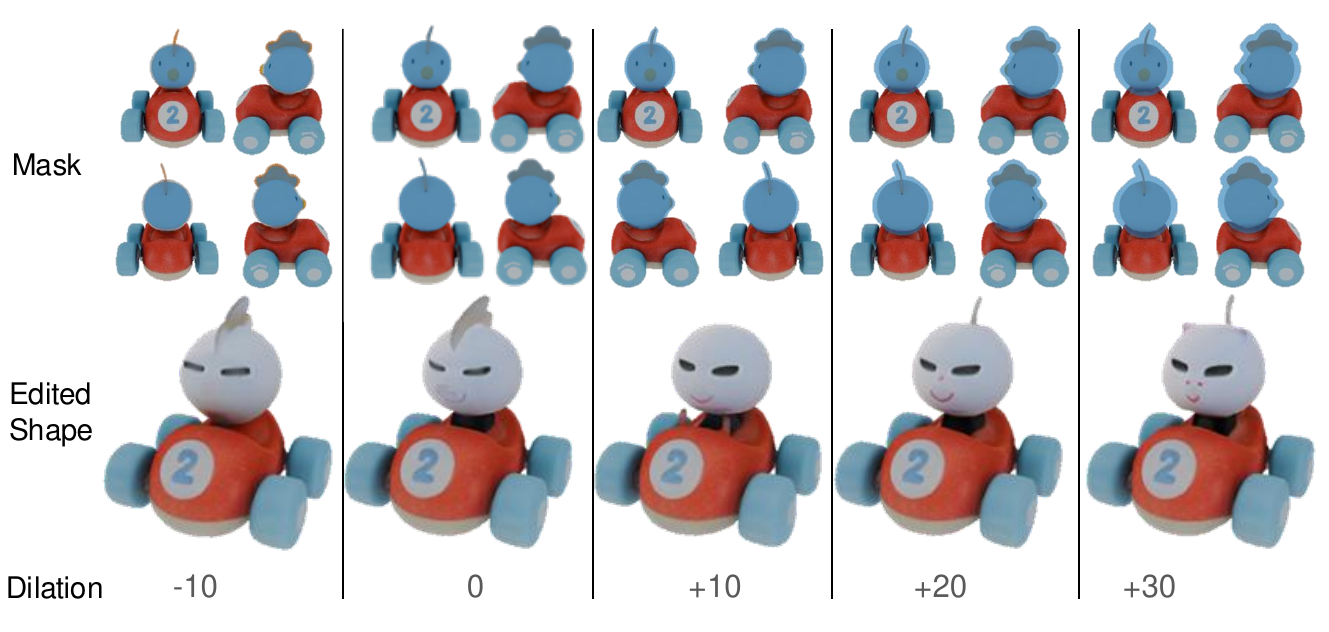}
    \caption{\textbf{Different granularity of masking}. Too fine-grained masks can over-constrain the generation process since they only point to the region to be replaced but do not include the user's intention. More dilation increases flexibility but can also edit more regions than intended (e.g., the region underneath the cat). Negative dilation means erosion.}
    \label{fig:mask_granularity}
\end{figure}

We experimented with different granularity levels for the input masks. We started with a mask that we detected automatically using Grounding DINO~\cite{groundingdino} and SAM 2~\cite{sam}. As shown in Fig.~\ref{fig:mask_granularity}. If we use the original segmentation, then the generation is restricted to that certain region and the model cannot have room to add "cat" features. That is, it tries to follow the shape of the original chicken. As we add more dilation, it tries to add features like cat ears. This shows the trade-off between loyalty to input and flexibility. Based on this observation, we gave coarse masks as input and allowed the model to edit flexibly. Thanks to our merging approach, we could still combine the edited region with the original shape to keep the rest intact.

\subsection{Directional CLIP Metrics}
\label{sec:supp_eval_metrics}

In Sec.~\ref{sec:eval}-\ref{sec:results}, we discuss directional CLIP score metrics~\cite{clip, gal2022stylegan, voxe} to evaluate 3D editing fidelity, to complement other quantitative metrics that measure the quality of the output shape. We report directional CLIP scores of different methods in Tab. 3 of the main paper. In this section, we formally define and discuss the reported metrics.

\begin{equation}
    \text{CLIP}_\text{dir} = \frac{1}{N}\sum_{i=1}^N<F^i_{IE} - F^i_{II}, F_{TE} - F_{TI}>,
\end{equation}
where $<., .>$ refers to an inner product, $F_{IE}^{i}$,  $F_{II}^{i}$ are the normalized CLIP image embeddings over rendered images of input and edited shapes, indexed by $i$, and $F_{TE}$, $F_{TI}$ are the corresponding normalized text embeddings of edited and input prompts. $i$ indexes a particular frame, while $N$ is the total number of rendered frames. In our directional CLIP evaluations, we use $N=70$ views rendered over a $360^\circ$ trajectory, significantly larger than the four input views we use for our method and the baseline methods.

We also introduce additional metrics inspired by $\text{CLIP}_\text{dir}$, but aim to fix some of its problems. First, we define 
\begin{equation}
\text{CLIP}_\text{dir-cos} = \frac{1}{N}\sum_{i=1}^N\text{C}(F^i_{IE} - F^i_{II}, F_{TE} - F_{TI}),
\end{equation}
where $\text{C}(., .)$ is the cosine distance.

We also introduce two modified versions of these metrics, namely 
\begin{equation}
\text{CLIP}_\text{dir-avg} = <\frac{1}{N}\sum_{i=1}^NF^i_{IE} - F^i_{II}, F_{TE} - F_{TI}>    
\end{equation}
\begin{equation}
\text{CLIP}_\text{dir-avg-cos} = \text{C}(\frac{1}{N}\sum_{i=1}^NF^i_{IE} - F^i_{II}, F_{TE} - F_{TI})
\end{equation}
that compute the same metrics over the average image embeddings instead of averaging scores to ensure further robustness.

We also propose two similarity change error metrics, $\text{CLIP}_\text{diff-edit}$ and $\text{CLIP}_\text{diff-noedit}$
\begin{equation}
    \text{CLIP}_\text{diff-edit} = \frac{1}{N}\sum_{i=1}^N|\text{C}(F^i_{II}, F_{TW}) - \text{C}(F^i_{IE}, F_{TW})|_\text{rel}
\end{equation}
\begin{equation}
    \text{CLIP}_\text{diff-noedit} = \frac{1}{N}\sum_{i=1}^N|\text{C}(F^i_{II}, F_{TG}) - \text{C}(F^i_{IE}, F_{TG})|_\text{rel}.
\end{equation}
Here, $|x - y|_\text{rel} = \frac{|x - y|}{\max(x, y)}$, $F_{TW}$ is the text embedding of the edited word or phrase, and $F_{TG}$ represents the "generic" text. For instance, when the prompt ``a chicken riding a bike" becomes ``cat riding a bike",  $F_{TW}$ embeds the text ``cat" and $F_{TG}$ embeds the text ``object riding a bike". By measuring similarity differences of rendered images to $F_{TW}$ and $F_{TG}$, we aim to measure the preservation of the object and context semantics, respectively.
